\documentclass{article}

\usepackage{microtype}
\usepackage{graphicx}
\usepackage{booktabs} 

\usepackage{amsmath}
\usepackage{amsfonts}
\usepackage{subcaption}
\usepackage[ruled,longend]{algorithm2e}
\usepackage[backref=page]{hyperref}

\usepackage[acronym]{glossaries}
\glsdisablehyper
\newglossaryentry{lcr}{
  name={lcr},
  description={Locally Constrained Representations},
  first={Locally Constrained Representations (LCR)},
  text={LCR},
}
\newacronym{mlp}{MLP}{Multi-layer Perceptron}
\newacronym{rl}{RL}{Reinforcement Learning}

\usepackage{hyperref}

\usepackage[accepted]{icml2024}

\usepackage{amsmath}
\usepackage{amssymb}
\usepackage{mathtools}
\usepackage{amsthm}
\usepackage{algorithmic}

\usepackage[capitalize,noabbrev]{cleveref}

\theoremstyle{plain}

\theoremstyle{definition}

\theoremstyle{remark}

\usepackage[textsize=tiny]{todonotes}

\icmltitlerunning{Locally Constrained Representations in Reinforcement Learning}

\begin{document}

\twocolumn[
\icmltitle{Locally Constrained Representations in Reinforcement Learning}



\icmlsetsymbol{equal}{*}

\begin{icmlauthorlist}
\icmlauthor{Somjit Nath}{mcgill,mila}
\icmlauthor{Rushiv Arora}{umass,dell}
\icmlauthor{Samira Ebrahimi Kahou}{mila,ets,cifar}
\end{icmlauthorlist}

\icmlaffiliation{mcgill}{McGill University}
\icmlaffiliation{umass}{University of Massachusetts, Amherst}
\icmlaffiliation{dell}{Dell Technologies, Research Office}
\icmlaffiliation{mila}{Mila-Quebec AI Institute}
\icmlaffiliation{ets}{École de technologie supérieure}
\icmlaffiliation{cifar}{CIFAR AI Chair}

\icmlcorrespondingauthor{Somjit Nath}{somjit.nath@mail.mcgill.ca}

\icmlkeywords{Machine Learning, ICML}

\vskip 0.3in
]



\printAffiliationsAndNotice{\icmlEqualContribution} 

\begin{abstract}
The success of \gls{rl} heavily relies on the ability to learn robust representations from the observations of the environment. In most cases, the representations learned purely by the reinforcement learning loss can differ vastly across states depending on how the value functions change. 
However, the representations learned need not be very specific to the task at hand.
Relying only on the \gls{rl} objective may yield representations that vary greatly across successive time steps. In addition, since the \gls{rl} loss has a changing target, the representations learned would depend on how good the current values/policies are. 
Thus, disentangling the representations from the main task would allow them to focus not only on the task-specific features but also the environment dynamics. 
To this end, we propose locally constrained representations, where an auxiliary loss forces the state representations to be predictable by the representations of the neighboring states. This encourages the representations to be driven not only by the value/policy learning but also by an additional loss that constrains the representations from over-fitting to the value loss.
We evaluate the proposed method on several known benchmarks and observe strong performance. Especially in continuous control tasks, our experiments show a significant performance improvement.
\end{abstract}

\section{Introduction}
Representation Learning is a crucial problem in machine learning, particularly in understanding how complex inputs can be represented in a compact and abstract manner while retaining useful information about the input. This has been studied more extensively for vision and natural language processing tasks compared to Reinforcement Learning (RL). RL has shown great success in games like Atari~\cite{mnih2015humanlevel}, Go~\cite{Silver2016,alphazero}, and Chess~\cite{alphazero}. However, its application in the real-world setting is limited due to challenges in representation learning.
In games, the input space is rather well-defined, e.g. it can be the state of the game for board games or it can be the image of the game screen in computer games \cite{mnih2015humanlevel}.

It has been shown that learning policies directly from high-dimensional inputs can be sample inefficient \cite{lake_ullman_tenenbaum_gershman_2017}. Additionally, in real-world scenarios, we often have inputs from different sensors that the agent needs to fuse into a compact representation, which can be challenging. 
Therefore, representation learning is a key to enabling RL agents to solve real-world problems.

Previous works on representation learning in \gls{rl} include~\cite{e2c} which uses supervised prediction as an auxiliary target, \cite{schwarzer2021dataefficient} which uses self-supervised learning, and \cite{rl_ae} which uses auto-encoders and therefore is completely unsupervised. In addition, \citeauthor{DBLP:journals/corr/JaderbergMCSLSK16} used auxiliary sub-tasks for representation learning. These methods use an auxiliary loss to learn the representation jointly. 

 

Here we introduce an auxiliary loss, where the objective encourages state representations to be predictable as a linear combination of the representations of \textit{nearby} states.

We structured our paper by first providing a brief background of the representation learning methods that exist in the literature (Sec. \ref{sec:back}). We then explain the proposed method and how it fits into the general landscape of representation learning methods for \gls{rl} in Sec.\ref{sec:alg}.
Following that, we detail the implementation of the algorithm and evaluate its performance on several benchmarks (Sec.\ref{sec:results}). Finally, we describe the key benefits of this algorithm and the scope of future improvements (Sec.\ref{sec:conclusion}).

\section{Background \& Related Works} \label{sec:back}
In \gls{rl}, the agent-environment interface is represented by Markov Decision Processes (MDPs), defined by $< \mathcal{S}, \mathcal{A}, \mathcal{R}, \mathcal{P} >$, where $\mathcal{S}$, $\mathcal{A}$, $\mathcal{R}$ and $\mathcal{P}$ represent the state space, the action space, the reward function and the transition function of the environment, respectively. RL is a sequential decision-making problem where the goal is to find an optimal policy (a mapping from states to actions). 

The key problem we intend to address here is how to find a good representation of the \textit{agent state} in conjunction with the main RL problem.
In this section, we will discuss the current approaches for representation learning and describe where our algorithm fits in the literature space.

\begin{figure*}
    \centering
    \includegraphics[width=\linewidth]{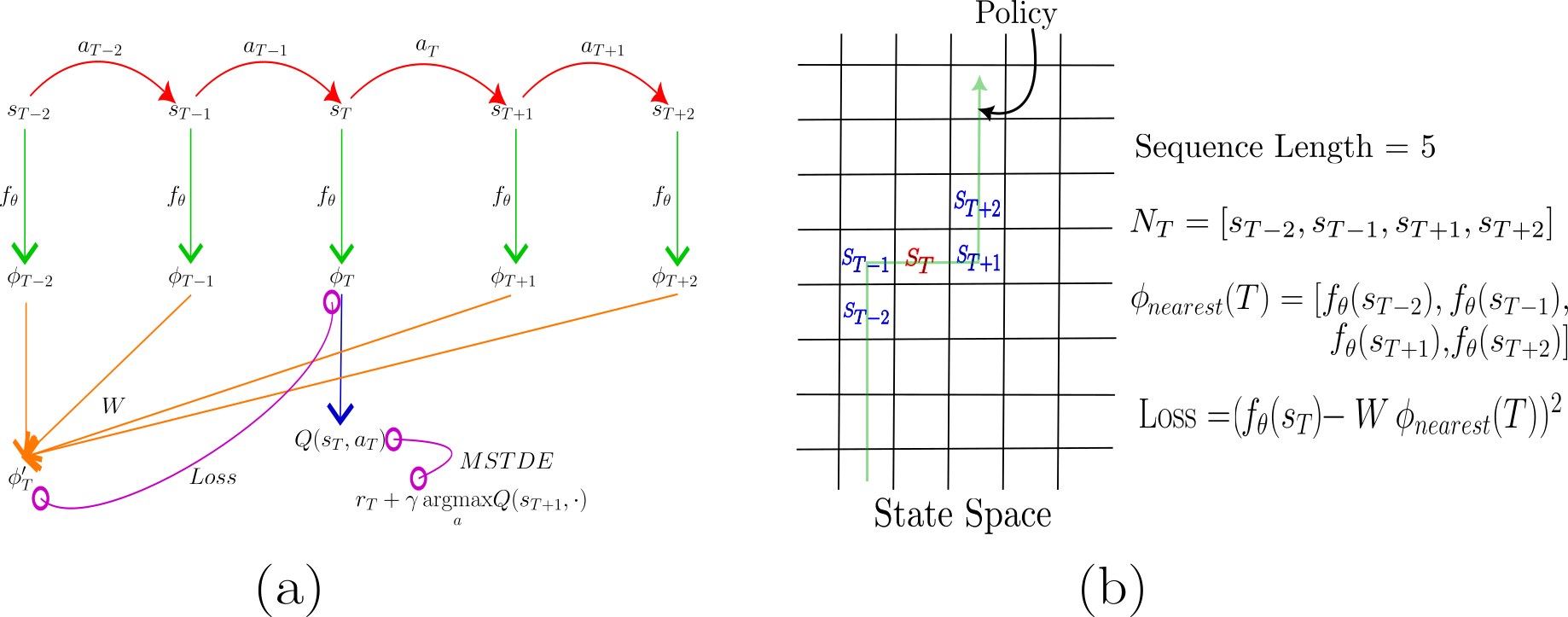}
    \caption{(a) We show the proposed training formalism of the Locally Constrained Representations Algorithm. We consider the time-step $T$, where at state $s_T$, the agent takes actions $a_T$. For value-based RL methods, we have Mean Squared TD Error (MSTDE) that learns the state values, $Q$. In addition, we have the \gls{lcr} loss, which is a Mean Squared Error (MSE) loss between the latent state $\Phi_T$ and a linear combination of the neighboring latent states, $\Phi_{T-2}$,$\Phi_{T-1}$,$\Phi_{T+1}$ and $\Phi_{T+2}$. This loss encourages the current representation $\Phi_T$ to be closer to this linear combination. (b) An example is shown for a sequence length of 5, where the current processed state is shown in red and the neighboring states considered are in blue. The loss calculated is fed directly to the \textit{Loss} in (a)}
    \label{fig:lcr}
\end{figure*}

\textbf{Auxiliary Losses:} Representation learning takes place by often minimizing two losses, the original \gls{rl} loss (which can vary depending on the type of \gls{rl} algorithm used and an \textit{Auxiliary} Loss. The auxiliary loss depends on the type of data being used. For example, it can be trying to predict the next reward or the next state or even reconstructing the current state~\cite{chung2018two}. It can also be General Value Functions \cite{10.5555/2031678.2031726} that can learn sub-tasks. The common approach is to learn a low-dimensional representation that is capable of reconstructing the current state. The commonly used method is Auto Encoders \cite{doi:10.1126/science.1127647} that can learn well from high-dimensional observations, such as images \cite{rl_ae, autoenc, e2c}. Recently, \cite{deepmdp} and \cite{schwarzer2021dataefficient} developed forward models that can predict the next state in the latent space without the need for reconstruction. These methods can predict multiple steps in the future instead of a one-step prediction.

The main problem with these approaches is that the targets are generally limited in the sense, that we only rely on data that the agent collects, if such data is not very informative then the entire method would be quite sub-par. For instance, predicting the next reward in a sparse-reward environment would yield very poor performance. Similarly, predicting the next few steps can also be really useful, but in environments with a lot of stochasticity such predictions can be poor and the loss would be trying to adapt to noise which can be an issue.


Most of the auxiliary losses mentioned in the previous paragraph are mean-squared errors with a particular variant of the target coming from the data. There exists a separate line of work involving data augmentation without the need for any additional loss. Data augmentation methods are commonly used for image input data. These methods randomly apply different transformations, such as cropping, rotation, or jitter, to generate additional training data, which helps the agent to be robust to small variations in the input space.
\citet{rad} and \citet{yarats2021image} showed that augmenting the states with such modulations can improve the performance of \gls{rl} without dedicated representation learning objectives. Although not strictly an auxiliary loss, we introduced this here to give context for contrastive learning~\cite{cpc,DBLP:journals/corr/abs-2002-05709} that takes advantage of similarity constraints between augmented data for representation learning. The most popular integration of contrastive learning comes with data augmentation, where positive pairs are modulations of the same image, whereas negative pairs are modulations from different images. The main goal of the algorithm \cite{DBLP:journals/corr/abs-2004-04136} is to increase the similarity between positive pairs while keeping negative pairs far apart. Other recent approaches include prototypical representations \cite{protorl}, predictive information \cite{predicitiveinfo}, and augmented temporal contrast \cite{atc}, \cite{tcn} to associate pairs within a short time. These methods attempt to make the representations of one state closer to the representations of neighboring states. 

\textbf{Constraining State Representations by Prior Knowledge:}
Often, the agents can leverage prior knowledge of the environment to constrain the representations. A very common method would be to assume that the ``interesting'' features of the state vary slowly, and thus setting a constraint on the state representation can be beneficial \cite{slowfeatures,pve}.
\citet{pve} added a prior that assumes the position of important objects vary and the algorithm enforces the representations to change accordingly.
Using both slow and variable features creates a trade-off and prevents the representations from becoming constant. \citet{roboticpriors} introduced two priors for robotics which are repeatability and proportionality. Repeatability introduces a constraint where the representations of two states would vary by the same amount for the same actions taken in the two states. 
Proportionality encourages the representation of a state to change by the same amount every time on multiple executions of the same action.

\glsfirst{lcr} algorithm adds a soft constraint on state representation using an auxiliary loss. Unlike \citet{roboticpriors}, \gls{lcr} does not require any prior knowledge of the environment. \Gls{lcr} assumes that \textit{the features of the state are linearly predictable by those neighboring it}. Thus, we do not explicitly have an additional target similar to the methods in the auxiliary losses. \gls{lcr} imposes the constraint that state representations should be linearly predictable by representations of surrounding states. From a theoretical perspective, it is similar to contrastive learning methods, except instead of enforcing a squashing function augmented version of states, we encourage linear predictability directly in the temporal space.

\section{Locally Constrained Representations} \label{sec:alg}
In this section, we describe the intuition behind this approach, formulate the algorithm, and provide details on its implementation.
\subsection{Intuition}
In most Reinforcement Learning environments, the main objects of interest change in a predictable manner governed by the dynamics of the environment. For example, in most domains, there is an underlying physics that acts on each object, for example, in MuJoCo~\cite{todorov2012mujoco}, gravity acts on the main agents and changes their positions in a well-defined manner.
Under this assumption, the dynamics should be relatively simple in local policy space, where the states are a few actions apart.

To exploit this for representation learning, we introduce the soft constraint that the representation of a state should be closely approximated by a linear combination of neighboring state representations.
This constraint encourages the state representation to remain close to a manifold on which local dynamics can be expressed in a simple function of linear form.
Without this constraint, the representations can change arbitrarily from one state to the next as the representations become intertwined with the local reward signal from the environment. This might result in over-fitting as the representation would fail to consider the dynamics of the environment.

Even predicting a single future step \cite{chung2018two} in a self-supervised learning loss can take advantage of the smoothness of the state space. Predicting multiple steps \cite{deepmdp, schwarzer2021dataefficient} in the future can improve the learned representation by helping it capture features that do not change much (for example: slow-changing features, such as objects).
However in \gls{lcr}, instead of doing prediction, we simply constrain the representation by the linear predictability assumption and since we use multiple neighboring states for the loss, it can also capture environment properties well. We test this hypothesis using a linear probe on a frozen representation in Appendix~\ref{sec:linear_probe}.

\Gls{lcr} is different from the contrastive learning approaches that attempt to pull together representations of neighboring states, as LCR merely introduces a linear explainability constraint to the representations. Thus, two neighboring representations can be far apart from each other, even if they are linearly related.

One important \textbf{limitation} of this approach is that, this method can only work well when this assumption is fully satisfied. For example, in domains like certain Atari games~\cite{bellemare2013arcade}, the screen can suddenly change depending on the actions, and in those scenarios, we do not expect our algorithm to work well.

\subsection{Algorithm}
The core idea of this method is to impose a constraint on the representation of a state and to encourage it to be linearly predictable by the representations of nearby states. This learning formulation is inspired by locally linear embeddings \cite{doi:10.1126/science.290.5500.2323}, where the embeddings are constrained to be linearly representable by the nearest data points.
Here we define the neighborhood to be a window of $K$ states in the sequence of transitions that are chosen by the current behavior policy. In this way, the representation of a state takes into consideration neighboring states instead of only focusing on the RL objective, which can improve generalization.
However, we do not want features that are completely unrelated to the main task, thus we only impose a \textit{soft constraint} on the representation.

\begin{algorithm}[h]
\caption{Locally Constrained Representations}
\label{alg:LCR}
\SetKwFunction{FMain}{\text{LCR}}
\SetKwProg{Fn}{\text{Function}}{:}{}
\Fn{\FMain{$steps, batch, sequence\,length=K+1$}}{
$t \leftarrow 0$ \\
\For{update in 1,2,3 $\dots$updates}{
    \text{Observe state}  $s$ \\
    \text{Take Action} $a$ \\
    \text{Train policy/value network by base RL algorithm} \\
    \If(\tcp*[h]{Run LCR on the last batch}){$t\, \% \,batch == 0$}
    {   
        $\text{All States} = GetLastBatch(t, batch)$ \\
        $\text{Initialize}\, W$ $\sim$ Uniform(0,1)\\
        $N \leftarrow$ \text Set of all Neighbors in Sequence \\
        \For{state in All States}
        {
            $N[T] = GetSequence(state, K+1)$ \\
        }
        $\phi_{nearest}(T) = f(N)$ \\
        Loss = $(f(\text{All States}) - W\phi_{nearest}(T))^2$ \\
        \For{$step\,\, in\,\, 1,2,3 \dots steps$}
        {
            \text{Run Gradient Descent on LCR Loss} \\
        }
    }
    }
}
\end{algorithm}
We need to keep a batch of state transitions of size $B$, based on which we constrain the representations. Since we experimented with online \gls{rl}, we need additional memory to keep track of the samples. If the main \gls{rl} algorithm already uses an experience replay buffer \cite{Lin2004SelfImprovingRA,mnih2015humanlevel}, we can directly take the last $B$ samples from the buffer. More training samples mean better state-space coverage and better generalization. So higher batch sizes are often better and this comes as a trade-off with respect to compute and memory requirements. 

Once we have a batch, $B$, of states, we obtain $K$ neighbors of each sample from the trajectory, assuming the total sequence length is $K+1$, including the sample itself.
We then add a constraint to encourage the representation of the sample to be predictable by the representations of the neighbors. We describe how this algorithm integrates with \gls{rl} in Algorithm~\ref{alg:LCR}.

Let $s_T$ be a sample drawn from the buffer of size $B$. Also, let $s_{T-K/2},\dots,s_{T-1},s_{T+1},\dots,s_{T+K/2}$ be the nearest neighbors.
Let 
$$\phi_T = f(s_T)$$
be the $D$-dimensional representation of state $s_T$, where $f$ is a parameterized function.
For all of our experiments, we use a neural network to implement $f$. 
Also, we define
$$\phi_{nearest}(T) = [\phi_{T-K/2},\dots,\phi_{T-1},\phi_{T+1},\dots,\phi_{T+K/2}]^\mathrm{T},$$
where $\phi_{nearest}(T)\in\mathbb{R}^{K \times D}$ is the matrix of neighboring state representations stacked row-wise.
We define the loss
$$\sum_{T=1}^{B}||W \phi_{nearest}(T)-\phi_{T}||_2^2, $$
where
$W \phi_{nearest}(T)$ is the predicted representation of the state by the linear combination of the neighbors with coefficients $W\in\mathbb{R}^{1 \times K}$. We learn $W$ and $f$ by minimizing this loss using gradient descent steps. After every new batch of samples of size $B$, this optimization is done in an interleaved manner with the standard \gls{rl} training. Since this is added as an auxiliary loss for representation learning, it can be applied on top of any \gls{rl} algorithm as an auxiliary loss.

We want a soft constraint on the representation and thus we do not solve this optimization completely but take only a few gradient descent steps while keeping $W$ positive (by clipping all negative weights to 0) to limit the solution space. This is because we are only solving this on a limited batch size with a short state-space coverage, so taking too many steps can overfit to the samples in the batch. This can also be a problem especially once the agent has stabilized, so we decrease the learning rate of \gls{lcr} as the algorithm learns as well, which prevents this over-fitting.
Figure~\ref{fig:lcr} (a) shows a training formulation of the method and Figure~\ref{fig:lcr} (b) demonstrates an example of the auxiliary \gls{lcr} loss.

\begin{figure*}[ht!]
\centering
\begin{subfigure}{.4\textwidth}
  \centering
  \includegraphics[width=\linewidth]{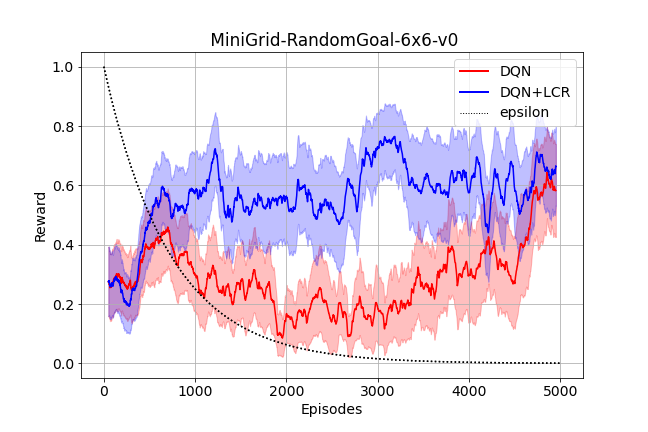}
\text{(a)}
\end{subfigure}%
\begin{subfigure}{.4\textwidth}
  \centering
  \includegraphics[width=\linewidth]{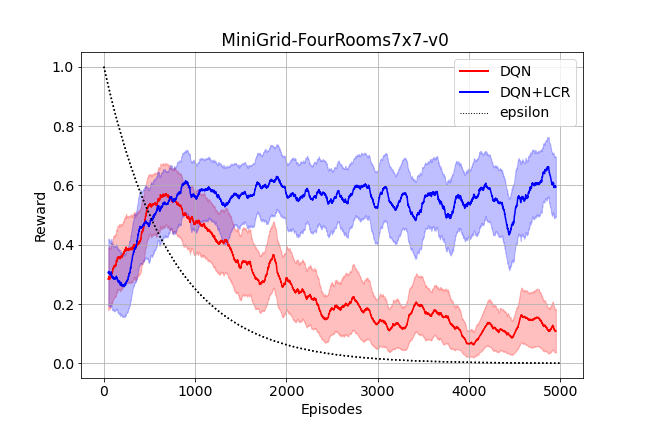}
\text{(b)}
\end{subfigure}
\caption{Performance of DQN (red) and DQN with LCR (blue) on the MiniGrid Environments. Both algorithms were trained for 10 runs with LCR using a sequence length of 11, 100 gradient steps, and a batch size of 5000. The detailed hyperparameters are mentioned in the appendix.}
\label{fig:minigrid}
\end{figure*}

\begin{figure*}[ht!]
\centering
\begin{subfigure}{.35\textwidth}
  \centering
  \includegraphics[width=\linewidth]{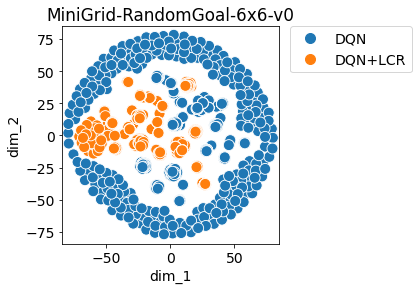}
\text{(a)}
\end{subfigure}%
\begin{subfigure}{.35\textwidth}
  \centering
  \includegraphics[width=\linewidth]{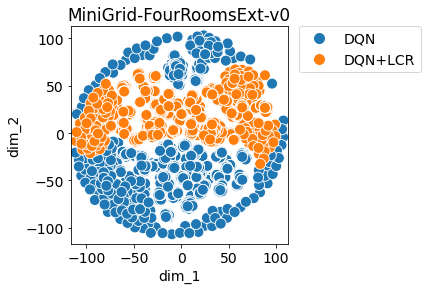}
\text{(b)}
\end{subfigure}
\caption{tSNE plots of the state representations obtained by 20 random trajectories of the respective environments. \gls{lcr} constrains the state representations by encouraging linear predictability with respect to its neighboring representations.}
\label{fig:tsne}
\end{figure*}
\section{Results} \label{sec:results}

\begin{figure*}[ht]
    \centering
    \includegraphics[width=\linewidth]{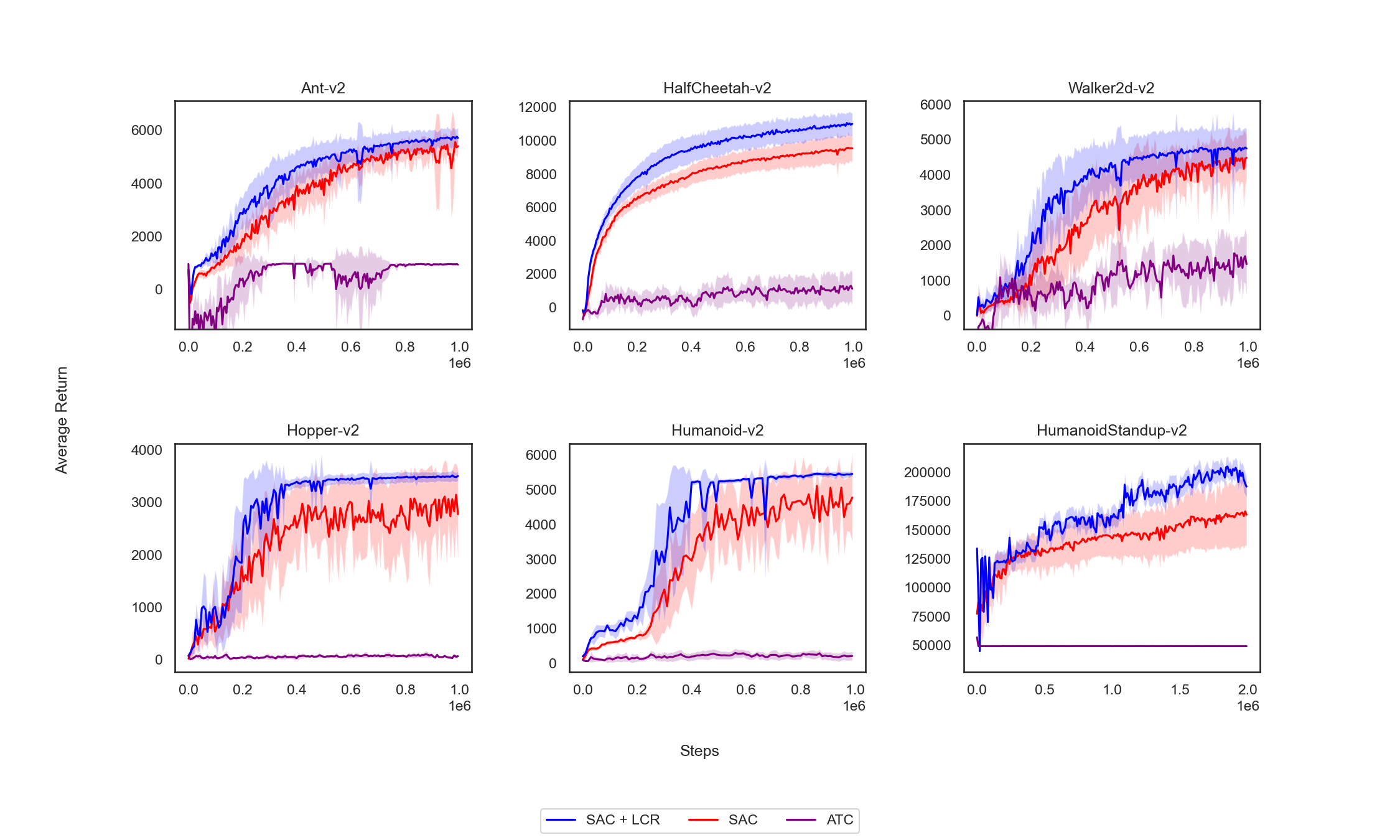}
    \caption{Training curves on 6 Mujoco environments using SAC with and without LCR across 10 runs. This figure highlights the impact of \gls{lcr} in these domains with well-defined physics. The poor performance of ATC in these settings highlights that constraining representations based on proximity is not a good strategy and adding linear predictability is much stronger.}
    \label{fig:mujoco}
\end{figure*}
\begin{figure*}[ht]
    \centering
    \includegraphics[width=\linewidth]{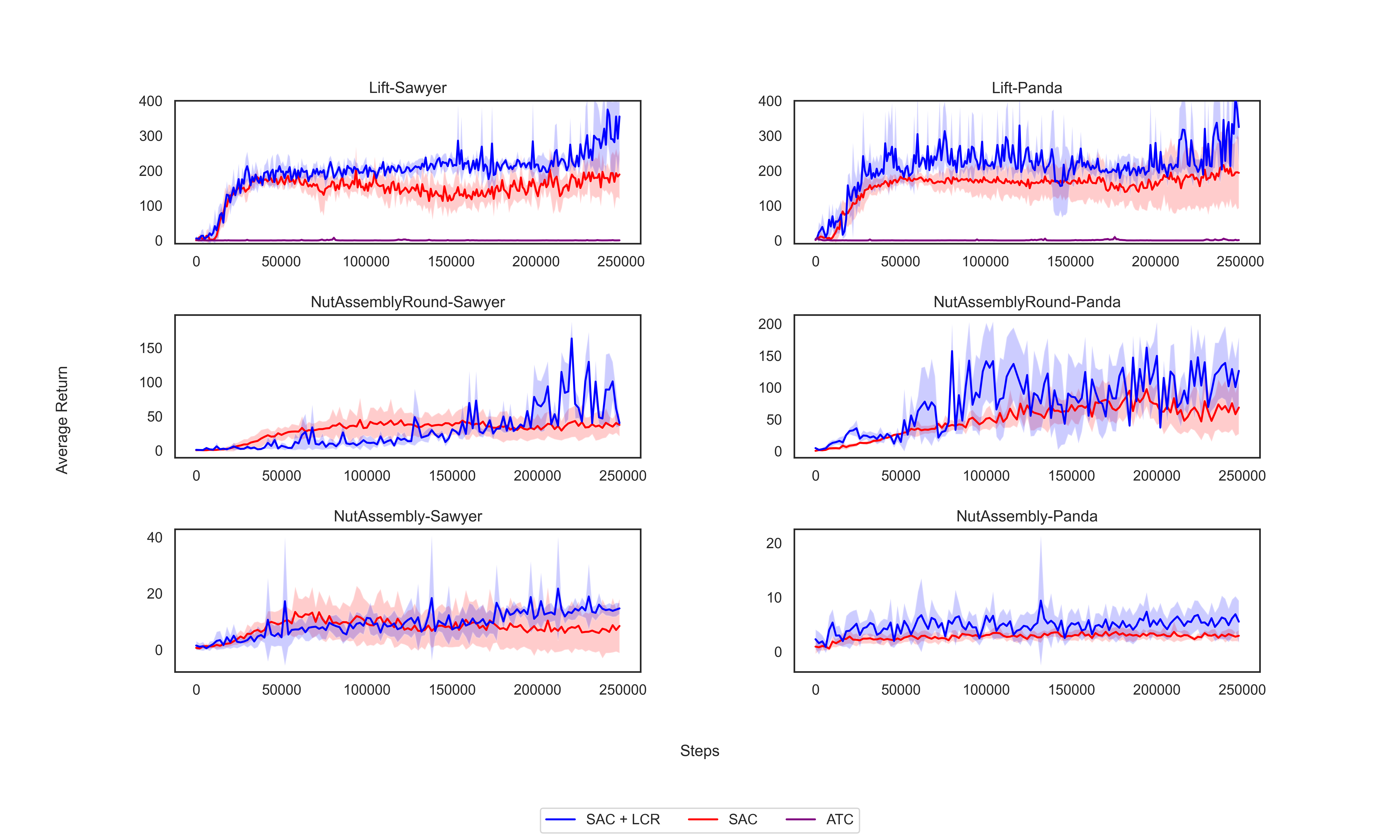}
    \caption{Training curves on 6 Robosuite environments with two robots Sawyer and Panda. We see that adding \gls{lcr} to the SAC consistently improves performance. All the curves are averaged across 10 independent seeds with the shaded portion representing the standard error.}
    \label{fig:robot}
\end{figure*}

In this section, we test our algorithm in 18 different environments (2 from MiniGrid environments \cite{gym_minigrid}, 6 environments from Mujoco \cite{todorov2012mujoco}, 6 environments from Robosuite and 4 environments from Arcade Learning Environment (Atari) \cite{bellemare2013arcade}). The reason for choosing most of these environments is that there are well-defined learnable dynamics. For diversity, we also analyze the performance of LCR in simpler environments such as CartPole and Acrobot in Appendix~\ref{sec:low}.

\subsection{Implementation Details and Baselines}
The description of the environments and the detailed hyper-parameters are mentioned in Appendix~\ref{sec:env} and~\ref{sec:add_exp} respectively. We provide ablation studies of the respective hyper-parameters introduced by \gls{lcr}, both in the main paper and in Appendix~\ref{sec:add_exp}.


For our baselines, we avoided comparisons to auxiliary representation learning methods which use data augmentation and are only designed for pixel space. We also exclude methods that have access to prior knowledge of the environment as \gls{lcr} does not need additional information. We mostly focus on environments with clear dynamics as \gls{lcr} is designed for this specific task.

We compare \gls{lcr} to a variant of Augmented Temporal Contrast (ATC)~\cite{atc}, a contrastive learning baseline for our experiments. We implemented a state-based version of ATC, without the augmentation. Our main aim in this paper is to highlight how this strategy of locally encouraging representations to be \emph{linearly predictable} can help in representation learning and our experiments henceforth highlight that.

\subsection{MiniGrid Environments}
\gls{lcr} helps in learning representations which prevents over-fitting in environments by constraining the representations locally. So, given a stochastic environment, is the representation learned by LCR more generalizable? We try to answer this question by choosing two environments from the MiniGrid suite~\cite{gym_minigrid}, where the goals change randomly after every episode. In both environments, the agent has to navigate a grid and reach a goal. The agent receives a reward of $+1$ for reaching the goal and $-0.01$ for each step that it takes. We designed the environment \emph{RandomGoal}, an extension of the EmptyRoom environment from MiniGrid~\cite{gym_minigrid} with the goal position changing randomly at the end of every episode. \emph{FourRooms} environment is the same as the generic one from MiniGrid~\cite{gym_minigrid}, but with reduced size to alleviate the problem of exploration which we are not studying in this paper. We run DQN~\cite{mnih2015humanlevel} and DQN with LCR in both environments with fully observable input. Since the output is a 2D matrix, both architectures have 3 convolutional layers, with the first 2 followed by a max-pooling layer. The output of the last convolutional layer is ``flattened'', i.e. reshaped into a vector. This entire network corresponds to $f_{\theta}$ in Fig. \ref{fig:lcr} (a). The output of this network is then passed to a value network which consists of two dense layers and which outputs the Q values. The entire flatten layer is the representation layer ($\phi$ in Fig. \ref{fig:lcr} (a)) on which LCR is applied. Exploration is handled by the decaying $\epsilon$-greedy strategy (dotted black line).

Figure~\ref{fig:minigrid} demonstrates that DQN+LCR performs better because it does not overfit to one goal location. This problem can happen with learning representations with the RL loss only. As a result, DQN (red) performs poorly in both settings, with complete failure in the FourRooms environment. Adding LCR on top of DQN however, improves the performance significantly, reaching near-optimal policies in both cases. Note, that the final rewards are noisy because the optimal reward changes with the location of the goal.

\textbf{Constrained Representations:}
Figure~\ref{fig:tsne} shows the tSNE~\cite{vanDerMaaten2008} plots in 2 dimensions for the representations learned by DQN and DQN with LCR after training for 5000 episodes. The plots represent the representations of the states $\phi$, in 20 trajectories with the same random policy (so that they all visit the same states equally). It can be seen that the spread of different state representations is much smaller for DQN+LCR even though both of the algorithms visit the exact same states. This indeed confirms that LCR learns a much more constrained representation. This is especially true for the \emph{RandomGoal} environment, where most of the states would be easily represented by neighboring states because the grid is empty. On the other hand, the representations learned by DQN without \gls{lcr} would be biased because value estimation is the only learning signal, and as a result, the representations are somewhat saturated which can lead to over-fitting.

\subsection{Mujoco and Robosuite Environments}
For our second set of experiments, we take \emph{Mujoco}~\cite{todorov2012mujoco, brockman2016openai} and Robosuite~\cite{robosuite2020} with well-defined physics, and observe the performance of \gls{lcr} in easy and difficult tasks. For these experiments, we used Soft Actor-Critic (SAC)~\cite{sac}, which is known to have a good performance on these tasks. We used the default architecture from \citet{sac} with 2 hidden layers followed by an output layer. The last hidden layer is used as the representation layer $\phi$ to which the \gls{lcr} loss is applied.
For \gls{lcr}, we used the following hyper-parameters: a batch size $B$ of 5000, a learning rate of $5 \times 10^{-4}$ with an exponential learning rate decay of 0.99, sequence length of 11 and 100 gradient steps for \gls{lcr} optimization. 

Figure~\ref{fig:mujoco} shows the learning curves on \emph{Mujoco}.
For relatively difficult tasks, such as Hopper-v2, Humanoid-v2, and HumanoidStandup-v2, we can see that adding LCR on top of SAC dramatically improves performance, and it learns a lot faster. This is probably because \gls{lcr} encourages the representations to capture the well-defined dynamics of Mujoco instead of just relying on the policy and value losses of SAC. It is interesting to note the poor performance of ATC~\cite{atc} in these domains. We used a modified version of ATC without augmentation to account for the state-based inputs. This highlights that using contrastive losses to squash representations together is too constrictive for representation learning, and we see that it negatively affects the performance of SAC.
Figure \ref{fig:robot} shows the learning curves for \emph{Robosuite}.
We evaluate the algorithms in a setting with 2 robots in 3 environments with varying degrees of difficulty.
Among the considered environments, to the best of our knowledge, SAC has only been shown to solve the \emph{Lift} environment. We note that SAC+LCR outperforms the baselines on the easy tasks while giving a significant learning advantage on the difficult ones.

The poor performance of contrastive learning suggests that just pushing local representations together is not a sound strategy as it constrains the representations in a manner that can hamper the performance of the main RL agent. This further lends credence to the assumptions of \gls{lcr}, which indicates that linear predictability is a better constraint on the feature space that prevents over-fitting.

\subsection{Atari}
\begin{figure*}[ht]
    \centering
    \includegraphics[width=\linewidth]{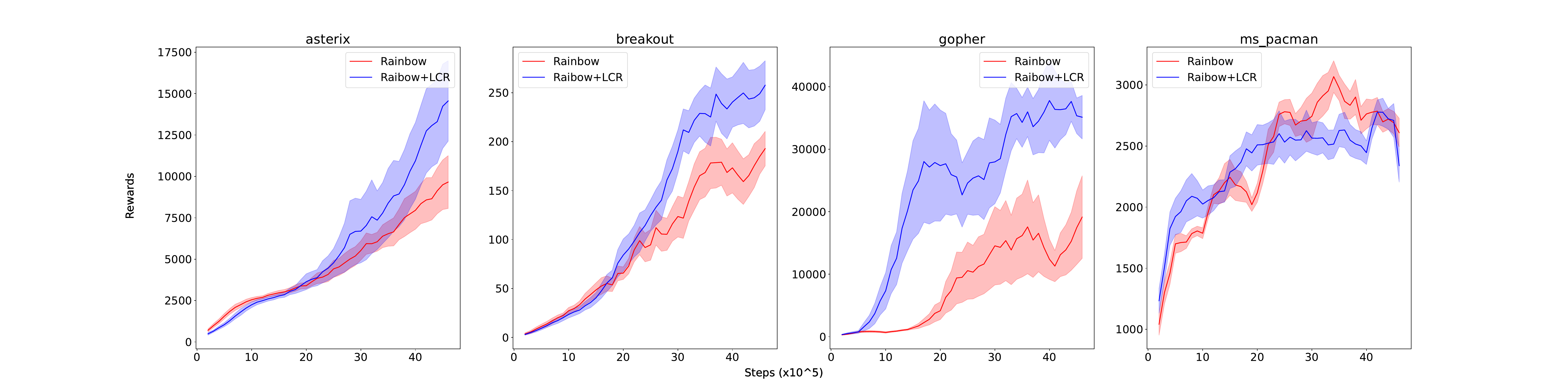}
    \caption{Training curves on 4 Atari environments using Rainbow with and without \gls{lcr} for 5 Million Frames across 5 runs. We choose games with well-defined physics and low stochasticity, and we observe that in these settings \gls{lcr} is more sample efficient.}
    \label{fig:atari}
\end{figure*}
\begin{figure*}[ht!]
\centering
\begin{subfigure}{.33\textwidth}
  \centering
  \includegraphics[width=0.7\linewidth]{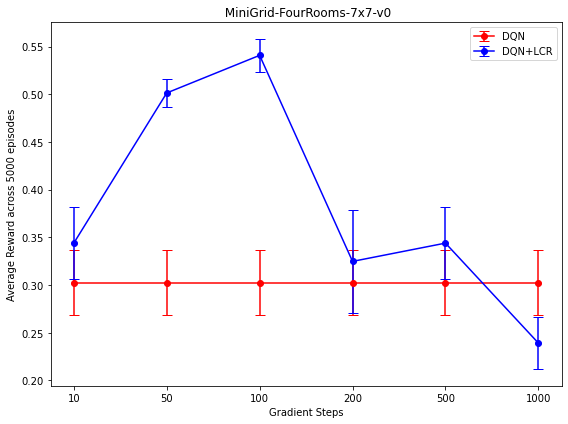}
\end{subfigure}%
\begin{subfigure}{.33\textwidth}
  \centering
  \includegraphics[width=0.7\linewidth]{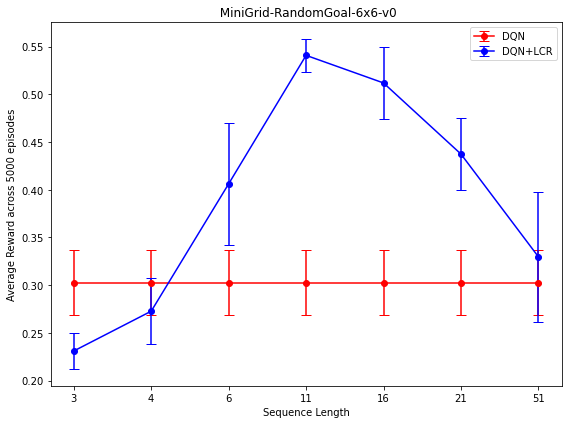}
\end{subfigure}
\begin{subfigure}{.33\textwidth}
  \centering
  \includegraphics[width=0.7\linewidth]{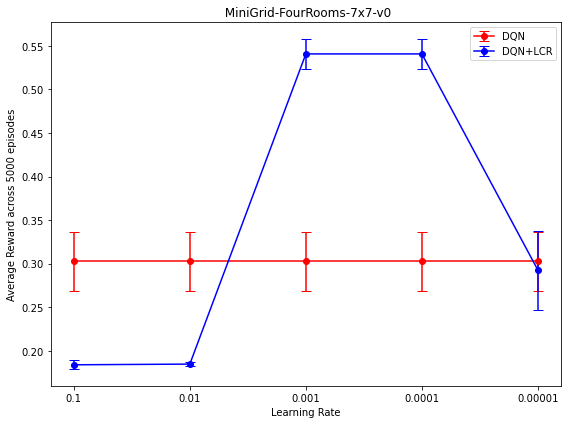}
\end{subfigure}
\caption{We study the sensitivity of performance with respect to \gls{lcr}'s hyper-parameters for the MiniGrid FourRooms environment over 10 runs. In each plot we vary a single hyper-parameter, keeping the other 2 at their default values (sequence length 11, 100 gradient steps, and learning rate $0.0001$).}
\label{fig:fr_param}
\end{figure*}


Although the \emph{Arcade Learning Environment} does not have continuous dynamics like Mujoco, we tested Rainbow~\cite{rainbow} with and without LCR in 4 Atari environments. For Rainbow, we used the original configuration of \citet{rainbow} with 3 convolutional layers followed by NoisyLinear layers~\cite{noisydqn}. The flattened output of the convolutional layer is chosen as the representation layer $\phi$ for \gls{lcr}.
We use a batch size of 1000 for \gls{lcr} because of the higher dimensionality of the state space (images).
To avoid over-fitting, we reduced the gradient steps to 20 and the learning rate to $1/10^{th}$ of the original RL learning rate.

Figure~\ref{fig:atari} shows the learning curves of \gls{lcr} compared with the baseline. Adding \gls{lcr} significantly improves the performance and sample efficiency in 3 out of 4 environments. In \emph{MsPacman} \gls{lcr} does not show improvement over the baseline but also does not seem to hurt.

\subsection{Ablation Studies}
LCR introduces 4 new hyper-parameters: the size of the batches, $B$ over which we apply LCR, the number of steps of gradient descent, the learning rate, and the neighborhood size $K$.
Figure~\ref{fig:fr_param}~(a) illustrates the parameter sensitivity on the \emph{MiniGrid} domains w.r.t. the number of gradient steps. In both environments, we find a similar pattern. With very few steps, LCR does not work well, and the same is true for a higher number of steps, where the representations are constrained heavily to focus more on the linearization of dynamics than the main RL task. Similarly, Figure~\ref{fig:fr_param}~(b) demonstrates the sensitivity to the size of the neighborhood. 
The optimal choice for this hyper-parameter depends on the environment. For example, in the \emph{RandomGoal} environment, we found larger neighborhoods do equally well if not better, because the grid is empty and having representations to be linearly represented by the neighbors does not affect performance. However, with larger neighborhoods in \emph{FourRooms}, \gls{lcr} does not perform as well because the representations are forced to be representable from a large portion of the state space, which violates the assumptions that neighbors are \textit{local}. Thus, we see a drop in performance for larger $K$. Figure~\ref{fig:fr_param}~(c) highlights the sensitivity to the learning rate. For most of our experiments, we found that too small of a learning rate does not help learning at all compared to DQN, because the updates to the representations are not big enough to have a sufficient effect on the actions taken. On the other hand, a too high learning rate can be detrimental, as the representations become too constrained. 
The fourth hyper-parameter, batch size, has no such trade-offs. A higher batch size reduces the chance of over-fitting on the main task, and results in better performance.

Detailed ablation studies across all the hyper-parameters are provided in Appendix~\ref{sec:add_exp}.

\section{Conclusion and Future Work} \label{sec:conclusion}
\gls{lcr} adds an auxiliary loss to constrain the state representations in a local policy space. This can improve the generalization and robustness of the representations learned. In any environment where the states do not change rapidly, with slow-moving features, LCR would show improvement. Furthermore, the addition of LCR does not hamper performance in simpler environments, where learning representations with the main RL loss is sufficient.

One potential weakness is that some environments may not be amenable to the linearization of dynamics imposed by LCR, resulting in degraded performance. In that case, the constraint may be relaxed by choosing a non-linear mapping. 
Another interesting future direction is to use a decoder to formulate the LCR loss via a downstream reconstruction loss.
The linear combination of neighboring representations would be decoded into a state reconstruction. The reconstruction loss can then be used for gradient computation of LCR.


\bibliography{ref}
\bibliographystyle{icml2024}

\newpage

\section*{Appendix}
\appendix
\section{Environments}\label{sec:env}
In this section, we describe the environments used in the main paper in detail.
\subsection{MiniGrid}
There are two MiniGrid environments used: RandomGoal and FourRooms. Both environments are based on~\citet{gym_minigrid}. For our experiments we used the fully observable state input.

\textbf{RandomGoal:} This environment is similar to the MiniGrid Empty Room environment, where the goal of the agent is to navigate a gridworld and reach a goal. The agent receives a reward of $+1$ on reaching the goal and also a negative reward of $-0.01$ per step it takes to reach that goal. The maximum steps per episode is $4\times\text{grid size}\times\text{grid size}$. The agent can take 3 actions: turn left, turn right, and go forward. The input state of the agent is provided in the form of a 2D matrix. After the end of every episode, the agent can spawn anywhere in the grid with the exception of the starting position of the agent.

\textbf{FourRooms:} This environment is an extension of the RandomGoal environment, except there are four rooms separated by walls. The reward structure, state dimension and the actions are the same. This is a little bit more challenging from a representation learning perspective, as the agent has to incorporate the collision with the walls.

\subsection{MuJoCo}
Multi Joint dynamics with Contact (MuJoCo)~\cite{todorov2012mujoco} is a physics engine supported by OpenAI Gym~\cite{brockman2016openai}. We used 8 environments from the MuJoCo suite.
\begin{enumerate}
    \item \textbf{Half-Cheetah-v2} is an environment where the RL agent is a two-legged robot. It needs to learn to sprint fast.
    \item \textbf{Ant-v2} is a similar environment, except that the agent controls a four-legged robot. The goal of the agent is to learn to sprint.
    \item \textbf{Humanoid-v2} is an environment in which the agent learns to control a bipedal robot to walk on the ground without falling over.
    \item \textbf{HumanoidStandUp-v2} is an environment, where the agent has to control a humanoid bipedal robot and get it to stand up.
    \item \textbf{Walker2d-v2} is an environment where the agent learns to operate a leg to hop forward.
    \item \textbf{Hopper-v2} operates an agent where the agent operates two legs that enable it to walk forward.
\end{enumerate}

\subsection{Robosuite}

The Robosuite \cite{robosuite2020} framework is a collection of robots and tasks powered by the MuJoCo physics engine. We experimented on three Robosuite environments of varying degrees of difficulty. We test on the following tasks:

\begin{enumerate}
    \item \textbf{Lift:} A single robot arm must lift the cube placed on the tabletop in front of it to a certain height. The initial cube location is randomized every episode.
    \item \textbf{NutAssembly:} A single robot arm works with colored nuts and pegs: it must place the square nut on the square peg and the circular nut on the circular peg.
    \item \textbf{NutAssemblyRound:} An easier version of \emph{NutAssembly} with only circular nuts and pegs. 
\end{enumerate}

On all three tasks, we experimented with both the Sawyer and Panda single arm robots.

\subsection{Atari}
We use the Arcade Learning Environment~\cite{bellemare2013arcade} to run our experiments on Atari 2600. The environments we used to test our algorithm are:
\begin{enumerate}
    \item \textbf{Asterix:} In this game, the agent needs to eat hamburgers while avoiding dynamites.
    \item \textbf{Breakout:} Breakout is a popular game, where the agent controls a paddle and the goal is to break all the bricks in the game using a ball.
    \item \textbf{Gopher:} In gopher, the agent is a person who needs to protect three carrots from a gopher.
    \item \textbf{Ms Pacman:} Here, the agent has to navigate a maze while eating pellets and avoiding ghosts.
\end{enumerate}

\subsection{Gym Control}
These are the classic control environments in Open AI Gym~\cite{brockman2016openai}. We used the popular CartPole-v1 and Acrobot-v1 environments.
\begin{enumerate}
    \item \textbf{Acrobot-v1:} In this environment, the agent controls a linked chain and the goal is to push the low-hanging chain up to a certain height while applying torque to the joint.
    \item \textbf{CartPole-v1:} The goal of this environment is to balance a paddle with a pole on it by applying forces to move the paddle to the left or right.
\end{enumerate}

\section{Verifying the quality of learned representations using a linear probe}\label{sec:linear_probe}
\begin{figure}[ht!]
\centering
\includegraphics[width=\linewidth]{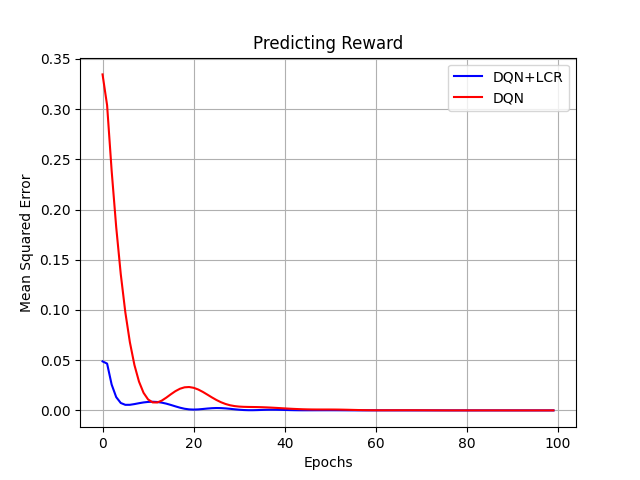}
\caption{Mean Squared Error loss with a linear probe on top of a frozen representation trained on MiniGrid-RandomGoal-6$\times$6 environment. The prediction target is the next reward. We notice faster learning with \gls{lcr}, suggesting the representations have captured the dynamics of the environment better. }
\label{fig:linear_probe}
\end{figure}
The goal of this experiment is to highlight that the representations learned by \gls{lcr} can capture information about the environment better than the base \gls{rl} algorithm. To test this hypothesis, we use a linear probe on a frozen representation to predict environmental dynamics. From Figure~\ref{fig:linear_probe}, we notice a substantial improvement in learning performance for predicting the next reward suggesting better representational quality.

\section{Adding LCR to low dimensional inputs}\label{sec:low}
\begin{figure*}[ht!]
\centering
\begin{subfigure}{.25\textwidth}
  \centering
  \includegraphics[width=\linewidth]{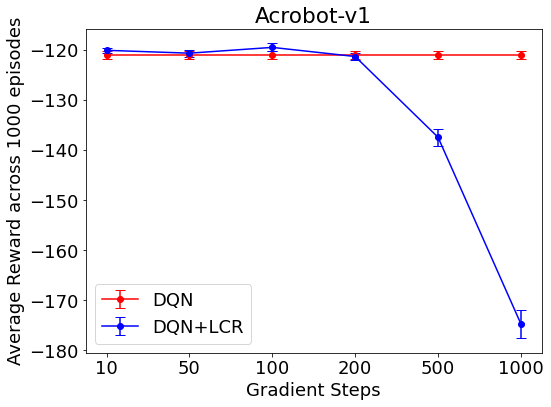}
\text{(a)}
\end{subfigure}%
\begin{subfigure}{.25\textwidth}
  \centering
  \includegraphics[width=\linewidth]{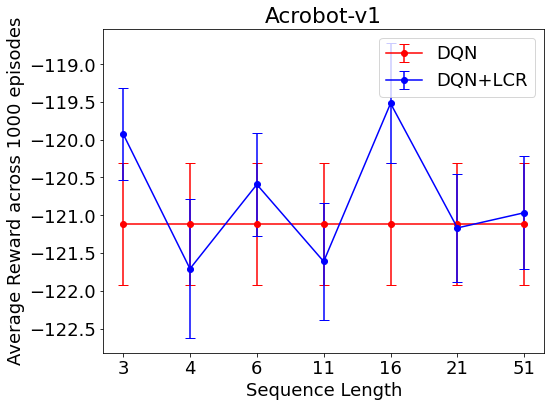}
\text{(b)}
\end{subfigure}
\begin{subfigure}{.25\textwidth}
  \centering
  \includegraphics[width=\linewidth]{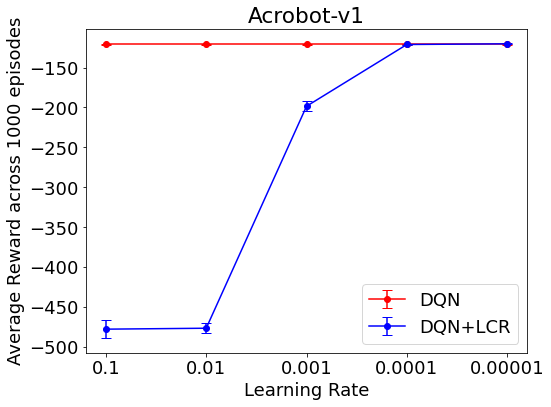}
\text{(c)}
\end{subfigure}
\begin{subfigure}{.24\textwidth}
  \centering
  \includegraphics[width=\linewidth]{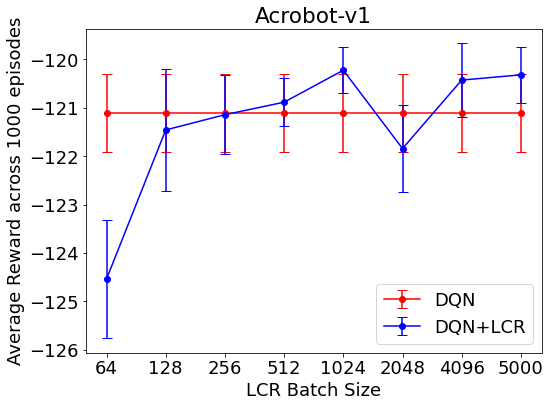}
\text{(d)}
\end{subfigure}%
\caption{Sensitivity to \gls{lcr} hyper-parameters for the Acrobot-v1 environment over 10 runs. The fixed hyper-parameters are a sequence length of 11, 100 gradient steps, and a learning rate of $0.0001$.}
\label{fig:acr_param}
\end{figure*}

\begin{figure*}[ht!]
\centering
\begin{subfigure}{.25\textwidth}
  \centering
  \includegraphics[width=\linewidth]{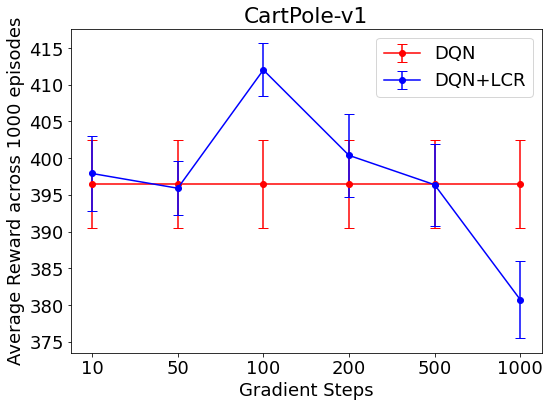}
\text{(a)}
\end{subfigure}%
\begin{subfigure}{.25\textwidth}
  \centering
  \includegraphics[width=\linewidth]{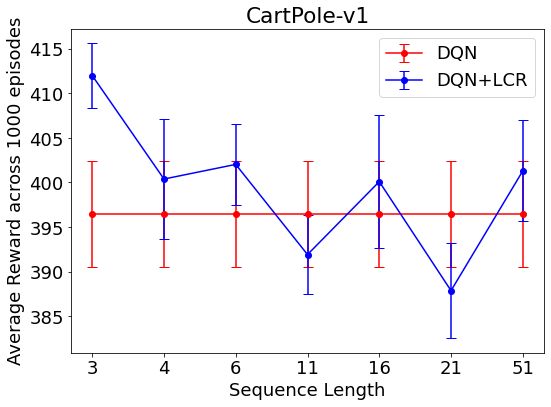}
\text{(b)}
\end{subfigure}
\begin{subfigure}{.25\textwidth}
  \centering
  \includegraphics[width=\linewidth]{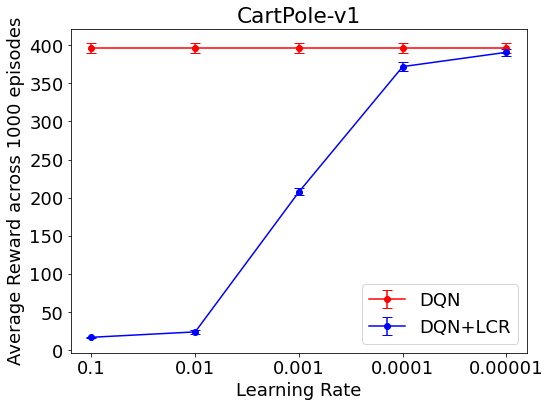}
\text{(c)}
\end{subfigure}
\begin{subfigure}{.24\textwidth}
  \centering
  \includegraphics[width=\linewidth]{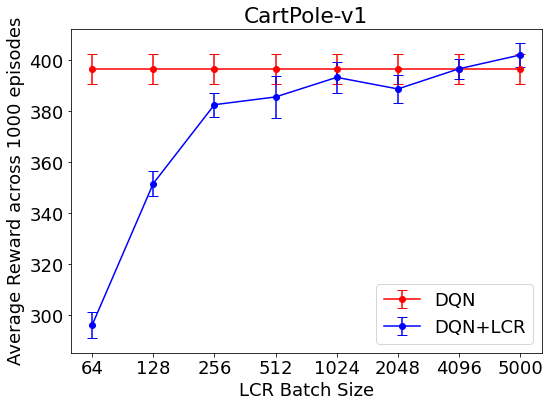}
\text{(d)}
\end{subfigure}%
\caption{Sensitivity to \gls{lcr} hyper-parameters for the CartPole-v1 environment over 10 runs. The fixed hyper-parameters are a sequence length of 11, 100 gradient steps, and a learning rate of $0.0001$}
\label{fig:cart_param}
\end{figure*}

In very simple \gls{rl} environments, separate representation learning is normally not required. Since we have the assumption of representations being linearly predictable by the neighbors, the representations for simple tasks might get constrained to the point that the agent does not learn anything. To test the performance of LCR with low-dimensional state space, we run LCR on two very simple environments, \emph{CartPole} and \emph{Acrobot} \cite{brockman2016openai}. For these environments we used simple DQN \cite{mnih2015humanlevel} with 2 hidden layers followed by the value layer. The output of the second hidden layer is the representation. From Figures~\ref{fig:acr_param} (b) and~\ref{fig:cart_param} (b), we can see that LCR does not impact the performance of the DQN over various sequence lengths. In fact, for some sequence lengths, DQN with LCR outperforms DQN.

\begin{figure*}[ht!]
\centering
\begin{subfigure}{.33\textwidth}
  \centering
  \includegraphics[width=\linewidth]{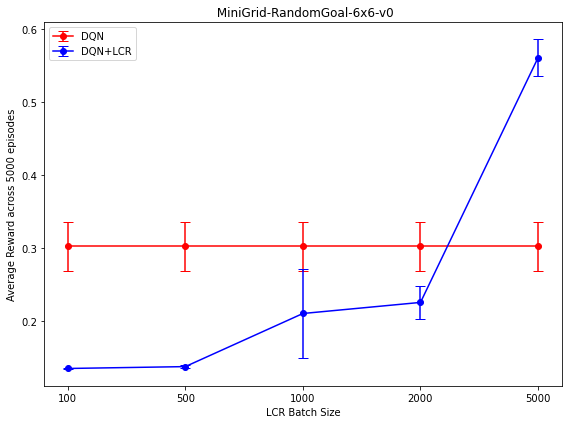}
\text{(a)}
\end{subfigure}%
\begin{subfigure}{.33\textwidth}
  \centering
  \includegraphics[width=\linewidth]{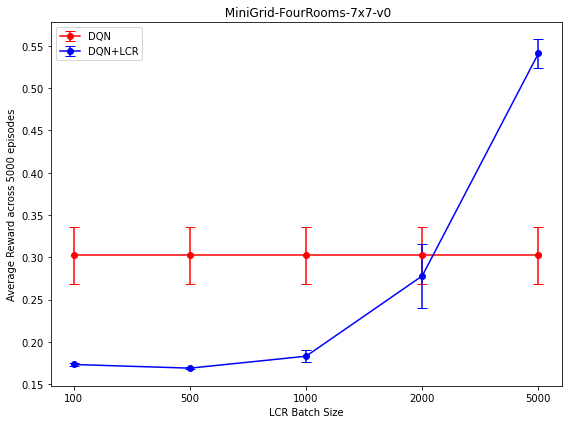}
\text{(b)}
\end{subfigure}
\caption{Sensitivity to the batch size of \gls{lcr}. The fixed hyper-parameters are a sequence length of 11, 100 gradient steps, and a learning rate of $0.0001$.}
\label{fig:batch}
\end{figure*}

\begin{figure*}[ht!]
\centering
\begin{subfigure}{.33\textwidth}
  \centering
  \includegraphics[width=\linewidth]{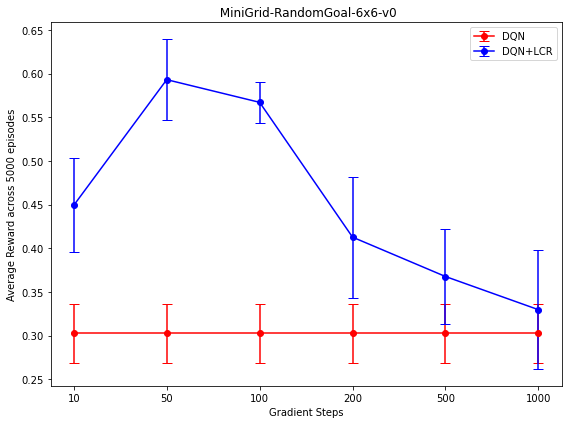}
\text{(a)}
\end{subfigure}%
\begin{subfigure}{.33\textwidth}
  \centering
  \includegraphics[width=\linewidth]{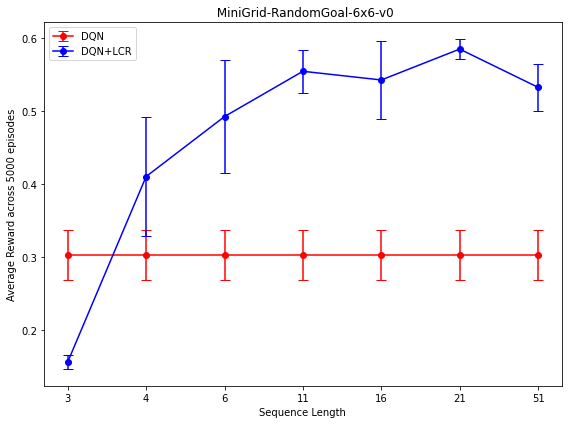}
\text{(b)}
\end{subfigure}
\begin{subfigure}{.33\textwidth}
  \centering
  \includegraphics[width=\linewidth]{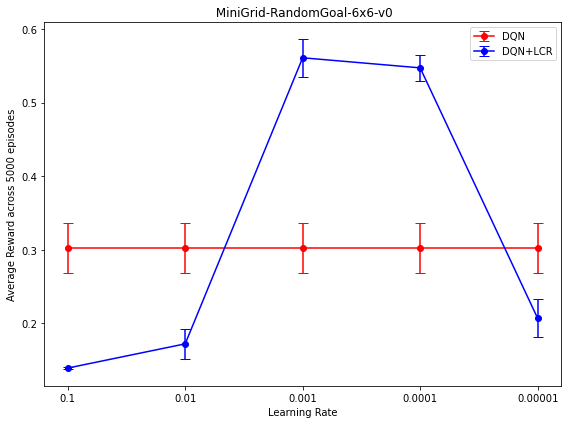}
\text{(c)}
\end{subfigure}%
\caption{Sensitivity to \gls{lcr} hyper-parameters for the MiniGrid RandomGoal environment over 10 runs. The fixed hyper-parameters are a sequence length of 11, 100 gradient steps, and a learning rate of $0.0001$.}
\label{fig:rg_param}
\end{figure*}

\section{Additional Experiments}\label{sec:add_exp}
\subsection{Sensitivity to Batch Size}
There is no trade-off for \Gls{lcr} optimization when it comes to batch sizes as higher batch size means more state space coverage and more global gradients per update. As demonstrated in Figure~\ref{fig:batch} a larger batch size is better in our experiments.
\subsection{Ablation studies for remaining environments}
In this section, we provide ablation studies on the remaining environments. Figure~\ref{fig:rg_param} are the sensitivity parameters with respect to the gradient steps (a), sequence length (b) \& learning rate (c), respectively. In Figure~\ref{fig:rg_param} (b), we see that the performance of LCR does not drop even for higher values of the sequence length. This is probably due to little variation in the states because it is an empty room and as a result, LCR can find a good solution. 

Figures~\ref{fig:acr_param} and~\ref{fig:cart_param} show the ablation studies for all the hyper-parameters in \emph{Acrobot-v1} and \emph{CartPole-v1} environments, respectively. Our observations in the MiniGrid experiments are also valid for these simple environments, i.e. locally linear representations do not hamper learning the original problem, provided the appropriate hyper-parameters are chosen.

The detailed hyper-parameters for all the algorithms are provided in Table~\ref{tab:my-table}.

\begin{table*}[!ht]
\centering
\caption{Hyper-Parameters of all experiments}
\label{tab:my-table}
\begin{tabular}{|l|l|l|l|l|}
\hline
Environments &
  Algorithm &
  Base Algorithm Parameters &
  LCR Parameters &
  Hardware and Software \\ \hline
MiniGrid &
  DQN &
  \begin{tabular}[c]{@{}l@{}}`runs' = 10\\ `episodes' = 5000\\ `batch\_size': 32,\\ `gamma': 0.99,\\ `learning\_rate': 1e-3,\\ `start\_epsilon': 1.0,\\ `stop\_epsilon': 1e-3,\\ `epsilon\_decay': 1e-3**(1/5000)\\ `hidden\_units': {[}64, 64{]},\\ `max\_buffer\_size': 10000,\\ `min\_buffer\_size': 1000,\\ `copy\_step': 5, \#episodes\end{tabular} &
  \begin{tabular}[c]{@{}l@{}}`K': 10,\\ `lcr\_batch\_size': 5000,\\ `gradient\_steps': 100,\\ `lcr\_learning\_rate': 1e-4\end{tabular} &
  \begin{tabular}[c]{@{}l@{}}Hardware-\\ CPU: Intel Gold 6148 Skylake\\ RAM: 6 GB\\ Software-\\ Tensorflow: 2.8.0\\ Python: 3.8\end{tabular} \\ \hline
MuJoCo &
  SAC &
  \begin{tabular}[c]{@{}l@{}}`runs' = 10\\ `max\_steps per episode':\\ `Ant-v2': 1e6,\\ `Walker2d-v2': 1e6,\\ `Hopper-v2': 1e6,\\ `HalfCheetah-v2': 1e6,\\ `Humanoid-v2': 1e6,\\ `HumanoidStandup-v2': 2e6,\\ $\alpha$=0.05 \\ Remaining hyper-parameters\\ same as \citet{sac}\end{tabular} &
  \begin{tabular}[c]{@{}l@{}}`K': 10,\\ `gradient\_steps': 100,\\ `lcr\_learning\_rate': 5e-4,\\ `lcr\_learning\_rate \\ exponential decay': 0.95,\\ `lcr\_batch\_size': 5000\end{tabular} &
  \begin{tabular}[c]{@{}l@{}}Hardware-\\ CPU: 2 Intel Gold 6240 \\ RAM: 192 GB\\ \\ Software-\\ Pytorch: 2.0.1\\ Python: 3.9\end{tabular} \\ \hline 
Robosuite &
  SAC &
  \begin{tabular}[c]{@{}l@{}}`runs' = 10\\ `max\_steps per episode':\\ 250000,\\ $\alpha$=0.05 \\ Remaining hyper-parameters\\ same as \citet{sac}\end{tabular} &
  \begin{tabular}[c]{@{}l@{}}`K': 10,\\ `gradient\_steps': 100,\\ `lcr\_learning\_rate': 5e-4,\\ `lcr\_learning\_rate \\ exponential decay': 0.95,\\ `lcr\_batch\_size': 5000\end{tabular} &
  \begin{tabular}[c]{@{}l@{}}Hardware-\\ CPU: 2 Intel Gold 6240 \\ RAM: 192 GB\\ \\ Software-\\ Pytorch: 2.0.1\\ Python: 3.9\end{tabular} \\ \hline
Atari &
  Rainbow &
  \begin{tabular}[c]{@{}l@{}}`runs' = 5\\ `frames' = 5 million\\ Remaining hyper-parameters\\ same as \citet{rainbow}\end{tabular} &
  \begin{tabular}[c]{@{}l@{}}`K': 10,\\ `gradient\_steps': 20,\\ `lcr\_learning\_rate': 6.25e-5,\\ `lcr\_batch\_size': 1000\end{tabular} &
  \begin{tabular}[c]{@{}l@{}}Hardware-\\ CPU: 6 Intel Gold 6148 Skylake\\ GPU: 1 NVidia V100\\ RAM: 32 GB\\ \\ Software-\\ Pytorch: 1.10.0\\ Python: 3.8\end{tabular} \\ \hline
Gym Control &
  DQN &
  \begin{tabular}[c]{@{}l@{}}`runs' = 10\\ `episodes' = 1000\\ `batch\_size': 64,\\ `gamma': 0.99,\\ `learning\_rate': 1e-3,\\ `start\_epsilon': 1.0,\\ `stop\_epsilon': 1e-3,\\ `epsilon\_decay': 1e-3,\\ `hidden\_units': {[}32{]}\\ `max\_buffer\_size': 5000,\\ `min\_buffer\_size': 100,\\ `copy\_step': 25, \#steps\end{tabular} &
  \begin{tabular}[c]{@{}l@{}}`K': 10,\\ `lcr\_batch\_size': 5000,\\ `gradient\_steps': 100,\\ `lcr\_learning\_rate': 1e-4\end{tabular} &
  \begin{tabular}[c]{@{}l@{}}Hardware-\\ CPU: Intel Gold 6148 Skylake\\ RAM: 1.2 GB\\ \\ Software-\\ Tensorflow: 2.8.0\\ Python: 3.8\end{tabular} \\ \hline
\end{tabular}
\end{table*}

\end{document}